# Statistical Modelling of Driving Scenarios in Road Traffic using Fleet Data of Production Vehicles

Christian Reichenbächer[1;2][0000-0002-0907-3287], Jochen Hipp[1][0000-0002-9037-9899] and Oliver Bringmann[2][0000-0002-1615-507X]

[1]Mercedes-Benz AG, Sindelfingen, Germany
[2]Department of Computer Science, University of Tübingen, Tübingen, Germany
christian.reichenbaecher@mercedes-benz.com, jochen.hipp@mercedes-benz.com, oliver.bringmann@uni-tuebingen.de

**Abstract.** Ensuring the safety of road vehicles at an acceptable level requires the absence of any unreasonable risk arising from all potential hazards linked to the intended automated driving function and its implementation. The assurance that there are no unreasonable risks stemming from hazardous behaviours associated to functional insufficiencies is denoted as safety of intended functionality (SOTIF), a concept outlined in the ISO 21448 standard. In this context, the acquisition of real driving data is considered essential for the verification and validation.

For this purpose, we are currently developing a method with which data collected representatively from production vehicles can be modelled into a knowledge-based system in the future. A system that represents the probabilities of occurrence of concrete driving scenarios over the statistical population of road traffic and makes them usable. The method includes the qualitative and quantitative abstraction of the drives recorded by the sensors in the vehicles, the possibility of subsequent wireless transmission of the abstracted data from the vehicles and the derivation of the distributions and correlations of scenario parameters.

This paper provides a summary of the research project and outlines its central idea. To this end, among other things, the needs for statistical information and data from road traffic are elaborated from ISO 21448, the current state of research is addressed, and methodical aspects are discussed.

**Keywords:** SOTIF, data-driven development, statistical representation.

## 1    Problem

An acceptable level of safety for road vehicles requires the absence of unreasonable risk caused by every hazard associated with the intended functionality and its implementation, including both hazards due to failures and due to insufficiencies of specification or performance insufficiencies [1, p. vi].

The "absence of unreasonable risk […] due to hazards […] caused by malfunctioning behaviour […] of E/E systems" is defined by ISO 26262-1 as "functional safety" [2, p. 14, with emphasis in the original]. For the achievement of functional safety, the standards of the ISO 26262 series define a process model. ISO 26262-3 [3] outlines procedures for performing a hazard analysis and risk assessment (HARA), to identify hazards at vehicle level and establish corresponding safety goals [1, p. vi]. The remaining standards [4], [5], [6], [7], [8], [9], [10], [11], [12] and [13] offer requirements and recommendations to avoid and control both random hardware failures and systematic failures that might compromise safety goals [1, p. vi].

According to ISO 21448 [1, p. vi], there are E/E systems that are free of the faults covered in the ISO 26262 series, but whose intended functionality and its implementation can still lead to hazardous behaviour. These can include, but are not limited to, systems based on sensing the external and internal vehicle environment in order to build situational awareness. ISO 21448 [1, p. vi] gives the following causes for potentially hazardous behaviour as an example, noting that these factors are especially relevant to functions, systems, and algorithms that use machine learning:

- the inability of the function to correctly perceive the environment;
- the lack of robustness of the function, system, or algorithm with respect to sensor input variations, heuristics used for fusion, or diverse environmental conditions;
- the unexpected behaviour due to decision making algorithm and/or divergent human expectations.

"The absence of unreasonable risk resulting from hazardous behaviours related to [such] functional insufficiencies is defined as the safety of the intended functionality (SOTIF)" by ISO 21448 [1, p. vi]. This aspect of safety is thus complementary to that addressed in the ISO 26262 series [1, p. vi]. In order to meet the requirements of ISO 21448, measures to eliminate hazards or reduce risks must be taken in the specification and design phase, the verification and validation phase as well as in the operation phase [1, p. vi].

Within the standard, each use case-relevant scenario is assigned to one of four areas (see Fig. 1) [1, Sec. 4.2.2]. Hazardous scenarios are those causing hazardous behaviour [1, Sec. 4.2.2].

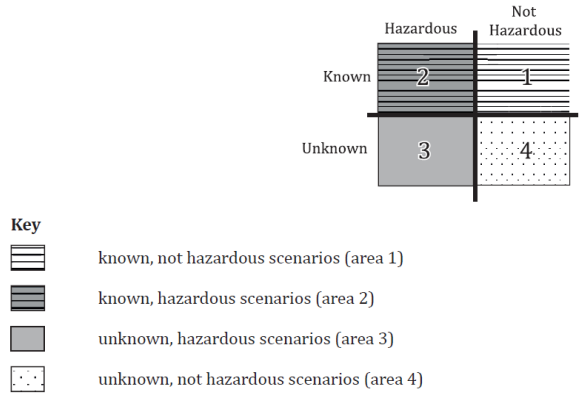

**Fig. 1.** Visualisation of scenario categories [1, Fig. 6]

Based on the conceptual abstraction of this model, the following goals of the SOTIF activities can be described:

— perform a risk acceptance evaluation of area 2 based on the analysis of the intended functionality;
— reduce the probability of known hazardous scenarios causing hazardous behaviour, in area 2, to an acceptable level through functional modification;
— reduce the probability of the unknown scenarios causing potentially hazardous behaviour, in area 3, to an acceptable criterion through an adequate verification and validation strategy [1, Sec. 4.2.2].

It's important to note that the size of the areas represents the number of scenarios, not the risk posed by those scenarios [1, Sec. 4.2.2]. To provide an argument for a sufficiently low risk of the intended functionality, not only the number of hazardous scenarios in these areas is relevant, but also the severity of the resulting harm and their probability of occurrence [1, Sec. 4.2.2].

"The ultimate goal of the SOTIF activities is to evaluate the potentially hazardous behaviour in hazardous scenarios and to provide an argument that the residual risk caused by these scenarios is sufficiently low" [1, Sec. 4.2.2]. For this purpose, the risk of known scenarios must be explicitly evaluated, while a sufficiently low risk from unknown scenarios must be proven by statistics-based testing [1, Sec. 4.2.2].

The sequence of activities required by ISO 21448 to achieve its goal is shown in Fig. 2. The circled numbers refer to the corresponding clauses in the standard, which go into more detail about the corresponding activities.

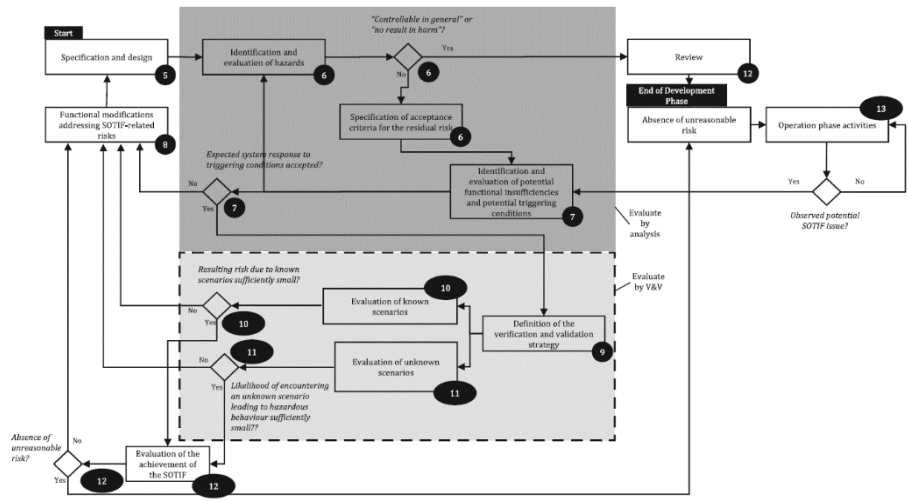

**Fig. 2.** Flow of ISO 21448 activities and corresponding clauses [1, Fig. 10]

In order to demonstrate compliance with ISO 21448, the objectives stated in the clauses must be achieved and proof of this must be documented in the corresponding work products [1, Sec. 4.3.2]. Several of those objectives require or at least recommend statistical information and other data about road traffic. In the following, these requirements and recommendations are briefly summarised. The key word "shall" indicates a requirement, the key word "can" a recommendation.

When it comes to the identification and evaluation of hazards, data shall be available for the derivation of acceptance criteria [1, Sec. 6.1]. Also relevant for the evaluation of hazards, there has to be distinguished between the occurrence of a triggering condition and the exposure to a scenario in which the hazard can lead to harm [1, Sec. 6.3]. Since the triggering conditions are in general not independent from scenarios, the statistical dependence between the probability of being in a scenario and the probability of encountering a triggering condition shall be taken into account, in order to use the exposure to a scenario within an argument for risk reduction [1, Sec. 6.3].

For the systematic analyses of potential functional insufficiencies and triggering conditions field experience can be considered [1, Sec. 7.3.1]. For example, when analysing requirements or accident data coming from records of the Data Storage System for Automated Driving / Event Data Recorder (DSSAD/EDR) [1, Tbl. 4]. However, in the authors' opinion, the inclusion of experience from the field also makes sense in the analysis of the ODD, use cases and scenarios as proposed by the standard [1, Tbl. 4].

The verification and validation strategy as well as validation targets "shall be defined and shall consider the sufficient coverage of the relevant scenario space" [1, Sec. 9.1]. The specification of the strategy can be derived using an appropriate combination of methods, including the analysis of field experience and lessons learnt, the use of databases with collected test cases and scenarios and the analysis of accident data [1, Sec. 9.3]. For each selected method, an appropriate development effort shall be defined and justified [1, Sec. 9.3]. Among other things, the distributions of the scenarios can be used for this purpose [1, Sec. 9.3].

When evaluating known scenarios, the residual risk from the hazardous ones is not unreasonable if their probability complies with the validation targets [1, Sec. 10.7]. From this it can be concluded that the probability of occurrence of these hazardous scenarios must be known.

The evaluation of unknown scenarios "shall demonstrate that the residual risk from unknown hazardous scenarios meets the acceptance criteria with sufficient confidence" [1, Sec. 11.1]. Note that one aspect of this is that the whole set of V&V activities shall cover the possible scenario space representatively [1, Sec. 11.1]. Different methods are provided for the evaluation of the residual risk arising from real-life situations, that could trigger hazardous behaviour [1, Sec. 11.3]. Amongst these ones: the search for unknown scenarios by systematically or randomly varying relevant scenario parameters to cover a diverse set of real-world scenarios.

Even after performing the activities in Clause 5 through 12 successfully, the risk evaluation may need to be reconsidered, if for instance previously unidentified hazards, functional insufficiencies and/or triggering conditions are uncovered or assumptions such as environment conditions or traffic regulation change [1, Sec. 13.2]. Therefore, before release, a field monitoring process shall be defined to be executed during the operation phase to maintain the SOTIF [1, Sec. 13.1]. For higher levels of driving automation, means such as Data Storage System for Automated Driving / Event Data Recorder (DSSAD/DER) may be required [1, Sec. 13.3]. For example, such an onboard monitoring mechanism can capture scenarios that triggered an emergency system reaction, where the driver unexpectedly took over and that lead to a minimal risk condition [1, Sec. 13.3].

It is self-explanatory that data on scenarios that trigger an emergency system reaction, where the driver unexpectedly takes over and that lead to a minimal risk condition can only be collected in the vehicle of the end users. As explained above, information is also required about the possible scenario space, about the statistical dependence between the probability of being in a scenario and the probability of encountering a triggering condition in it as well as about the probability of occurrence of known hazardous scenarios. Assuming that sufficient representativeness cannot be ensured by isolated, stationary traffic observation, only methods that enable data collection in the end user's vehicle prove to be relevant for the acquisition of this knowledge as well. In order to capture, represent and make knowledge usable in this way, the methods used for knowledge modelling must meet special limitations, including with regard to the available sensor technology, the storage capacity, the bandwidth for potential wireless data transfer and, last but not least, the protection of personal data.

## 2 Objective of the Research Project

The research project presented here is intended to provide an answer to the question of how representative data collected by production vehicles can be modelled into a knowledge-based system that represents the probabilities of occurrence of the concrete driving scenarios across the statistical population of road traffic and makes them usable. The working hypothesis to be (partially) tested to answer this question includes the following steps in order:

1. abstraction of the drives recorded by the sensors in the vehicles
2. wireless transmission of the abstracted data from the vehicles
3. derivation of the distributions and correlations of scenario parameters

Investigations into knowledge acquisition, such as the technical implementation of data collection in the vehicle and the wireless transmission of abstracted data from the vehicle, are not part of the research project. Instead, the project focuses on the abstraction of drives recorded by the sensors in production vehicles and the derivation of the distribution and correlations of scenario parameters.

## 3    State of Research and Reference to Theory

The modelling of driving scenarios has already been the subject of various scientific papers. Geyer et al. [14] present in their article a fundamental ontology with a metaphorical terminology. Bach et al. [15] take up the terminology for their domain model, which enables a description of real driving scenarios down to the logical level.

King et al. [16] use the domain model of Bach et al. [15] and present an approach for the identification of individual driving manoeuvres and logical driving scenarios in virtual test drives. The identification is carried out by means of knowledge-based metamodels of driving scenarios, driving data abstracted according to the same model and pattern recognition. The StreetWise method also relies on the use of knowledge-based metamodels for pattern recognition to identify logical scenarios and revises the ontology of Elrofai et al. [17], Geyer et al. [14] and Ulbrich et. al [18] [19]. De Gelder et al. [20] presents another, revised ontology for driving scenarios. In addition to the theoretical elaborations, an method for pattern recognition with its application to real data was described by Reichenbächer et al. [21].

Erdogan et al. [22] compare three different methods, one rule-based, one supervised, and one unsupervised, to extract driving manoeuvres from real-world driving data. Montanari et al. [23] propose an automated extraction of sequential, parameterized manoeuvres in a format such as OpenSCENARIO XML. In doing so, they enable the variation of manoeuvre parameters. Weisser [24] shares the idea of using an adaptable re-simulation based on manoeuvres and developed, implemented, verified and validated an automated process for the conversion of driving scenarios from field to OpenSCENARIO XML, respectively re-simulation. His approach allows to preserve the value of the state variables like spatial position and velocity at the end of each action, and thus improves the accuracy of the re-simulation.

The presented methods of knowledge representation use formal languages and notations for the representation of knowledge. Formal languages are abstract languages in which, in contrast to natural languages, the focus is usually not on communication, but on the definition and application of formal systems in the narrower sense and logic in the broader, general sense. Formal languages, however, are not intended to represent the common probability distribution of all the variables involved, as is necessary to solve the problem described.

De Gelder et al. [25] already present a metric based on waterstone distance, which can be used to quantify the representativeness of parameter sets generated for testing over real-world driving scenarios. To this end, they also extract parameters to describe driving scenarios, but also determine the probability density function of the parameters.

How a knowledge representation for driving scenarios, for example, an ontology, can be extended to include a common probability distribution of all variables involved is an aspect of the field of science that has not yet been sufficiently researched.

## 4 Method

For the experimental design and verification of our approach, recordings of the bus signals of production vehicles are available. In addition, a high-definition map provides information about the static environment of the drives.

For the modelling of driving scenarios, a distinction is made between qualitative and quantitative abstraction. Qualitative abstraction provides a linguistic description of the driving event. The quantitative abstraction builds on the qualitative abstraction and supplements concrete information with which the driving events can be reconstructed accurately.

### 4.1 Qualitative Abstraction of Drives

The qualitative and quantitative abstraction is explained in this article by means of a temporal excerpt of a drive recorded by a production vehicle. Fig. 3 schematically shows the dynamic driving behaviour of the three road users involved during this temporal excerpt. The ego vehicle in red is initially following the vehicle coloured in green on the right lane. Both vehicles do not change lane during the recorded time span and keep their velocity approximately constant. Another vehicle depicted, in brown is overtaking the ego vehicle on the left lane, changes to right lane afterwards and decelerates. After a certain time, this brown vehicle is accelerating again, changes back to left lane and overtakes the green vehicle.

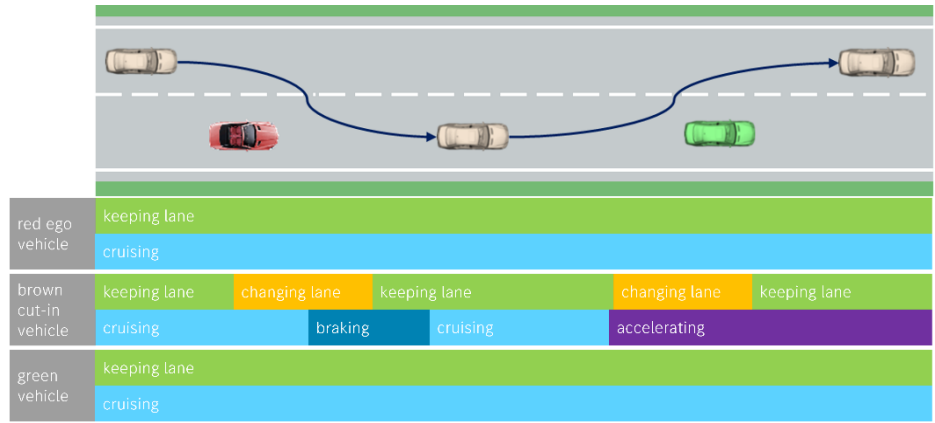

**Fig. 3.** Exemplary drive and its qualitative abstraction

The qualitative abstraction of the bus signals or the driving process is carried out by means of universal logical elements of an ontology, which is represented by a domain model, as presented in [21]. The approach makes use of the lane coordinate system (s/t) as defined in [26]:

> To every lane specified in a lane section of a road [...], there is a s/t-type coordinate system assigned. The lane's center line defines the s-axis going in the middle between lane's side boundaries throughout the whole lane section in the direction of the road's s-axis. The shape of the s-axis line is determined by the geometry of the respective lane. The s-axis lies on the road surface and therefore takes into account an elevation of the road (its inclination in the s-direction). The origin of s-coordinates is fixed to the beginning of the lane section.
>
> In contrast, multiple t-axes can be defined along the s-axis. Each t-axis points orthogonally leftwards to the s-axis direction and originates on the s-axis at the point with the concerned s-coordinate. All t-axes lie on the surface of the road and therefore adopt a lateral slope profile and an elevation of the road. [...]

Fig. 4 illustrates the process of qualitative abstraction of dynamic behaviour using the brown cut-in vehicle. The left diagram shows the temporal derivation of the s-coordinate over time, the right diagram the t-coordinate also over time. When the centre of a vehicle crosses a lane marking, the vehicle is assigned to the new lane. This also changes the reference coordinate system for the t-coordinate, which explains the jumps in the right diagram.

Each vehicle should be assigned exactly one lateral and one longitudinal action at any given time. This is done through automated recognition and instantiation of the respective action-elements of the universal ontology. In Fig. 4, this assignment is represented by the vertical black bars. Fig. 3 shows the assignment of actions for all vehicles involved in the drive recorded.

An advantage of this approach: logical scenarios that are predefined by the universal logical elements can get identified in the driving data by a pattern recognition method. In this way, for example, cut-in scenarios can be automatically detected, in which the cut-in vehicle decelerates immediately after changing lanes. The realisation of such a pattern recognition method is described in detail in [21].

On the basis of a purely qualitative abstraction, however, only a limited statement can be made about the criticality of a scenario. This is because there is a lack of information, such as the speed and distances between the road users involved. For this reason, abstraction is extended by a quantitative level.

### 4.2 Quantitative Abstraction of Drives

The quantification of abstraction has already been in introduced in [24] and is done by parameterizing the instances of the universal element classes. To do this, each logical element of the ontology is assigned a model equation that describes the change of the relevant state variable. This model equation can be, for example, a 5th degree polynomial (see (1)).

$$P(x) = a_0 x^5 + a_1 x^4 + a_2 x^3 + a_3 x^2 + a_4 x + a_5 \tag{1}$$

In this case, x represents the time, $P(x)$ the s- or t-coordinate and six degrees of freedom are available with the six coefficients ($a_0$ to $a_5$) to adapt the parameter equation to the measured data. Fig. 4 shows in blue how the dynamic behaviour of the cut-in vehicle can be abstracted with 5th degree polynomial equations. One was assigned to each instanced action, except for the lane change actions. The latter were abstracted with two parameter equations each, one for the period before and one for the period after the change of the lane assignment.

When fitting in the parameter values for the velocity, it should be ensured that their integral doesn't differ from the measured distance covered over the respective period. This is the only way to prevent larger deviations accumulated by the actions over longer abstraction periods. In addition, the parameter equations should be continuous at their edges. In this example, the remaining degrees of freedom were selected for the lowest possible square error.

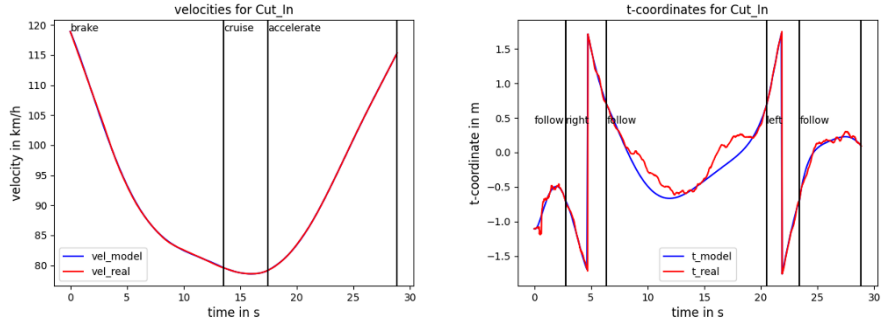

**Fig. 4.** Abstraction of the dynamic behaviour of the brown cut-in vehicle

It is still an open research question which type of parameter equation is optimal for which logical action. Conceivable are purely data-driven approaches, such as the presented polynomial equation, but also approaches with semantic parameters. An approach how it is used in the example at hand also requires intermediate storage of signals over a period at least equal to the duration of the action to be abstracted. It is possible, however, that the abstraction can be further subdivided within an action to reduce the intermediate memory requirement, similar to when changing lanes. If necessary, an abstraction can also be calculated continuously or a few points, for example, start, line crossing and end, are sufficient to describe an action such as a lane change with sufficient precision. Other aspects to be considered include the extent to which the parameter values of a model equation should be correlated with each other in order to enable subsequent statistical evaluation.

The quantitative description of the recorded drives can be automatically converted into the scenario description language OpenSCENARIO XML for simulation. The implementation of this has been described in [24]. This enables closed-loop re-simulation of individual driving scenarios as well as re-simulation of entire test drives. The focus of the research project described at hand is on providing information on the dynamic behaviour of road users. For this reason, the various simulation options, such as highly precise environmental modelling that would also support the sensor simulation, are not investigated or discussed in greater detail at this point.

### 4.3 Formal representation of correlations and distributions

The method used to derive and represent the correlation and distribution of parameters, actions and other elements of an ontology, as presented here, is still an open research question. The methodology with which the authors will address this problem involves a prioritized theoretical comparison as well as the selection of potentially suitable approaches and subsequent experimental implementation and further investigation of these.

In the theoretical part of the methodology, the combination of an ontology with a Bayesian network to map the common probability distribution of all variables involved will be examined in more detail. An experimental implementation can be carried out, for example, by analysing the shape of lane changes as a function of various parameters. Parameters that could be taken into account in the concept study are, among other things, the scenario of which the lane change is a component, the longitudinal speed with which the lane change is performed, as well as the parameters of the lane change model equation itself.